\documentclass[conference]{IEEEtran}
\usepackage{cite}
\usepackage{amsmath}
\usepackage{amssymb}
\usepackage{booktabs}
\usepackage{fixltx2e}
\usepackage{url}
\usepackage{hyperref}
\usepackage{graphicx, multicol, verbatim}
\usepackage{caption}
\usepackage{subcaption}

\begin{document}
%
\title{Lessons Learned Developing an Assembly System for WRS 2020 Assembly Challenge}

\author{\IEEEauthorblockN{Aayush Naik\IEEEauthorrefmark{1},
Priyam Parashar,
Jiaming Hu and
Henrik I. Christensen}
\IEEEauthorblockA{\IEEEauthorrefmark{1}anaik@ucsd.edu}
\IEEEauthorblockA{UC San Diego}}


%


\maketitle
\thispagestyle{plain}
\pagestyle{plain}

\begin{abstract}
The World Robot Summit (WRS) 2020 Assembly Challenge is designed to allow teams to demonstrate how one can build flexible, robust systems for assembly of machined objects. We present our approach to assembly based on integration of machine vision, robust planning and execution using behavior trees and a hierarchy of recovery strategies to ensure robust operation. Our system was selected for the WRS 2020 Assembly Challenge finals based on robust performance in the qualifying rounds. We present the systems approach adopted for the challenge.
\end{abstract}


%
\IEEEpeerreviewmaketitle

\section{Introduction}\label{sec:intro}

\begin{figure}
\centering
\includegraphics[width=0.48\textwidth]{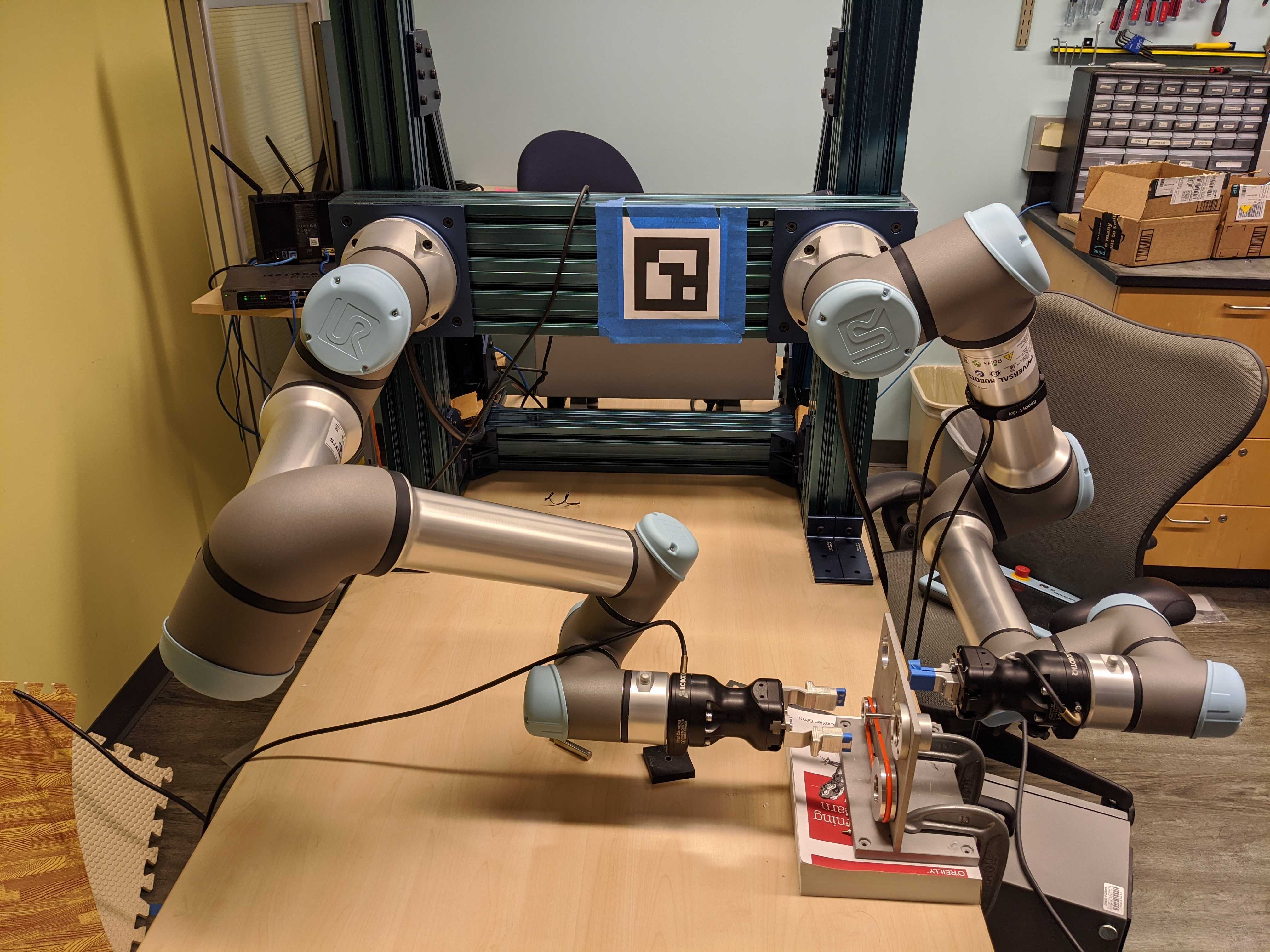}
\caption{Our robots working on the WRC 2020 taskboard.}
\label{fig:robot-photo}
\end{figure}

The World Robotics Summit 2020 Assembly Challenge, also known as World Robot Competition 2020 (WRC 2020), was designed to explore approaches to build robot systems that provide robust performance for high-mix/low-volume assembly production. How can we design a system that allows for flexible programming of an assembly problem and provides the needed robustness for operation in an environment with a minimum of fixtures, varying light conditions, and rapid changes in tasks/components? The competition involves a nominal setup for assembly (as shown in figure \ref{fig:robot-photo}) and a surprise change that is similar to the nominal setup. The system must be able to adopt for the surprise change in minimum time. During WRC 2018, the rubber belt was changed to be a metallic chain (i.e., no elasticity). This paper describes the design of a system for WRC 2020, which has three layers of abstraction - mission, task, and behavior. The basic behaviors are composed from a taxonomy of capabilities \cite{huckaby2014knowledge}. Behaviors are composed into tasks using behavior trees as the unifying framework. Robustness is achieved through use of behavior tree decorators and use of visual inspection. Missions are generated using an hierarchical task network \cite{nau1999shop} planning framework that directly takes failures and variations into account.

The main contributions of this paper are a description of our system, our approach to various problems/challenges we faced, and the lessons. We designed a framework which integrates deliberative planning with lower-level planners and sensors, uses a behavior tree to instantiate the process as a structure that allows for reactive state-checking and recovery, and supports overall system health with a watchdog that monitors processes. 

In the next section we talk about related work on flexibility in robot systems. In section \ref{sec:wrc}, we talk about WRC 2020 in more depth, highlighting the technical challenges in the competition. In section \ref{sec:approach}, we describe our approach. In section \ref{sec:impl}, we walk through an example execution, going through the underlying implementation. We also describe our codebase details. In section \ref{sec:perf}, we talk about the performance of our system and mention some obstacles we faced. Finally, in section \ref{sec:lesson} we highlight some lessons learned developing an assembly system for WRC 2020.







\section{Related Work}\label{sec:rw}



Most industrial robots for assembly or manufacturing employ the fixed “teach and playback” method. Consequently, these systems lack flexibility, both, in terms of agility (ability to quickly perform a changeover for a new product) and leanness (reusability of components, both software and hardware) \cite{wrs2020rules}. Highly flexible manufacturing/assembly has been the domain of job shops. Although many approximate computational solutions exist \cite{hoitomt1993practical, applegate1991computational, zhang1995reinforcement} to make the hard problem of job scheduling more tractable, job shops still remain limited in efficiency because of high product variability.

More recently, with the work of Jorg et. al \cite{jorg2000flexible}, who use a multi-sensor approach for an insertion task, and Lopez-Juarez et. al, who use learning from perception to adapt changing environments, \cite{lopez2005design} we see examples of a move towards flexibility. Over the last decade, international competitions have attracted a lot of attention towards flexibility in robot systems. The Amazon Picking Challenge (APC) (2015 thru 2017) is a great example of this. The participating teams integrated state-of-the-art perception and planning in robot systems. The goal for the first challenge (2015) was to design an autonomous robot to pick items from a warehouse shelf and place it into a bin. In APC 2015, ``some of the teams reported that they developed too many components from scratch and did not have time to make them robust'', and other teams noted that ``off-the-shelf software components they used as `black-boxes'  hid  important functionality that could not be properly customized.'' \cite{correll2016analysis} Thus, we make judicious use of standard open-source software packages (see section \ref{sec:std-pkgs}), but also tweak and repackage them with our codebase to avoid the blackbox problem. We also note that many teams employ intelligent hardware design and embodiment to reduce the complexity of their system \cite{eppner2016lessons}.
WRC 2018 was the first iteration of the WRC. It consisted of a kitting task (similar to APC) and an assembly task. APC is limited to first order constraints of picking and placing the right objects in the right location. WRC Assembly has additional second order constraints: specific fits/attachments between two (or more) parts. For WRC 2018, team O2AS \cite{von2020team} developed a system with multiple robots to manage the taskspace more efficiently. WRC 2018’s winning team from SDU introduced the concept of ``robot cell matrices'', which are reconfigurable and adaptable units of production or assembly, as part of their implementation \cite{schlette2020towards}. They rely heavily on task-specific solutions like holders and fixtures, and rapid manufacturing techniques like 3D printing to cope with changes in requirements.
Some common software tools that we have seen used repeatedly to improve flexibility of robot systems are Behavior Trees, motion plannning and control frameworks/libraries like MoveIt \cite{coleman2014reducing}, Orocos \cite{bruyninckx2001open} and OMPL \cite{sucan2012open}, and general purpose vision systems. Behavior Trees originated from the video game industry, but were popularized in robot systems by \"Ogren et al. \cite{ogren2012increasing} \cite{colledanchise2018behavior}.  A noteworthy example, the Navigation2 project \cite{macenski2020marathon2}, which is the ``spiritual successor'' of the widely used ROS Navigation Stack, uses behavior trees for ``navigator task orchestration''. We use such software tools along with lessons learned from previous works to develop our assembly system.

\section{WRC 2020}\label{sec:wrc}

\begin{figure}
    \centering
    \includegraphics[width=0.48\textwidth]{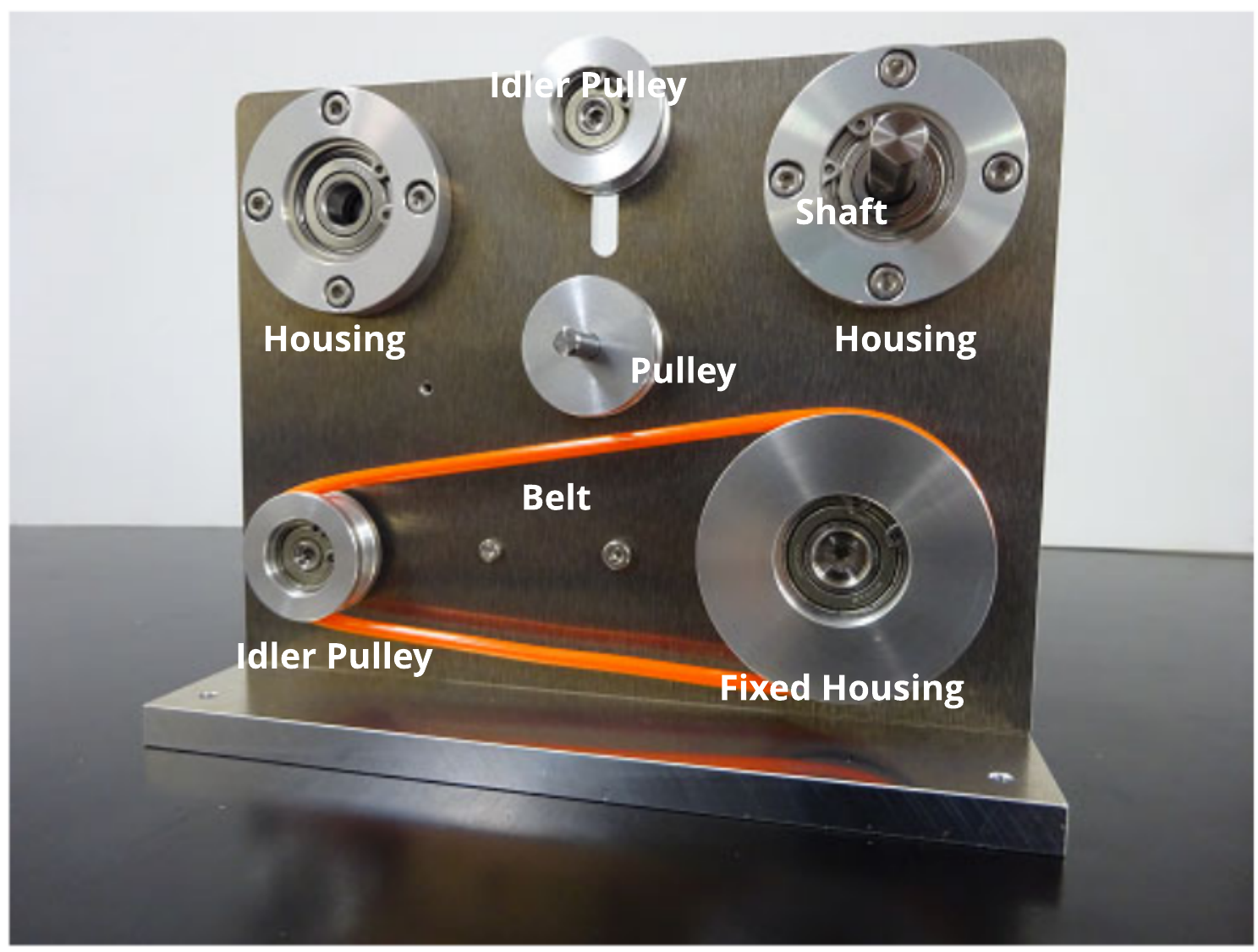}
    \caption{Assembled taskboard from WRC 2020 rulebook \cite{wrs2020rules}.}
    \label{fig:taskboard}
\end{figure}

To spur developments towards increasingly autonomous production/assembly systems, the World Robot Summit 2020 Assembly Challenge (WRC 2020) ``aims at realizing future manufacturing systems that can respond to variously changing orders (ultimately, even an order for a one-off product) by reconfiguring the system in an agile and lean manner.'' \cite{wrs2020rules}  
See figure \ref{fig:taskboard} for example taskboard assembly. The following factors make WRC 2020 a challenge:

\subsubsection{Tight allowances} One of the components of the assembly task is inserting a shaft into a hole in a housing. This is a classic peg-in-the-hole problem. However, the allowance is very tight; it is a transition fit that is difficult even for humans to do correctly (we tried and failed 8 out of 10 times). This can also lead to the shaft jamming into the hole, which would require manual intervention to fix.

\subsubsection{Small, featureless objects} Many parts used for assembly in WRC 2020 have high specular reflection, radial anisotropy, and are relatively featureless. Some parts, like nuts, are very small in size. This makes it difficult to accurately estimate the pose of these objects with traditional methods or in the absence of extensive 3D modelling of objects.

\subsubsection{Unknown apriori location of objects} The parts used for assembly will be delivered to the robots in carts delivered by automated guided vehicles (AGVs). Thus, we cannot make any assumptions about apriori location of objects and must rely on our vision system to detect the objects accurately.

\subsubsection{Surprise products} On day 3 of the competition, participants will be provided with a surprise product specification. The participants must be able to adapt their system for a new product in a short period of time.

\subsubsection{Hard and soft objects} Most objects in the competition are hard and metallic. There is also a soft belt that must be looped over the grooves of two pulleys. Thus, the participants must be able to deal with both kinds of objects.

\section{Our Approach}\label{sec:approach}

We organize our assembly system into three levels of abstraction: mission (highest), task (middle), and behavior level (lowest). The mission and task level decomposes the overall assembly plan into a series of tasks (represented as a behavior tree). It also does failure recovery and replanning in case of mission-level failures. The behavior level contains definitions and programs for execution of various skills like moving the robot arm, opening/closing grippers, and insertion. The behavior tree for each task is composed of these skills as action nodes. There is also a transparent ``system'' level which consists of hardware (robots and sensors), networking, operating system, ROS \cite{quigley2009ros} and a process orchestrator like Supervisor\footnote{http://supervisord.org/index.html}. The system level is responsible for recovery from non-planning based failures in a transparent way.

In the following sections we describe our hardware setup, camera setup, task planning, execution framework, and software packages. We also briefly talk about our vision system.

\subsection{Hardware Setup}

\begin{figure}
    \centering
    \includegraphics[width=0.48\textwidth]{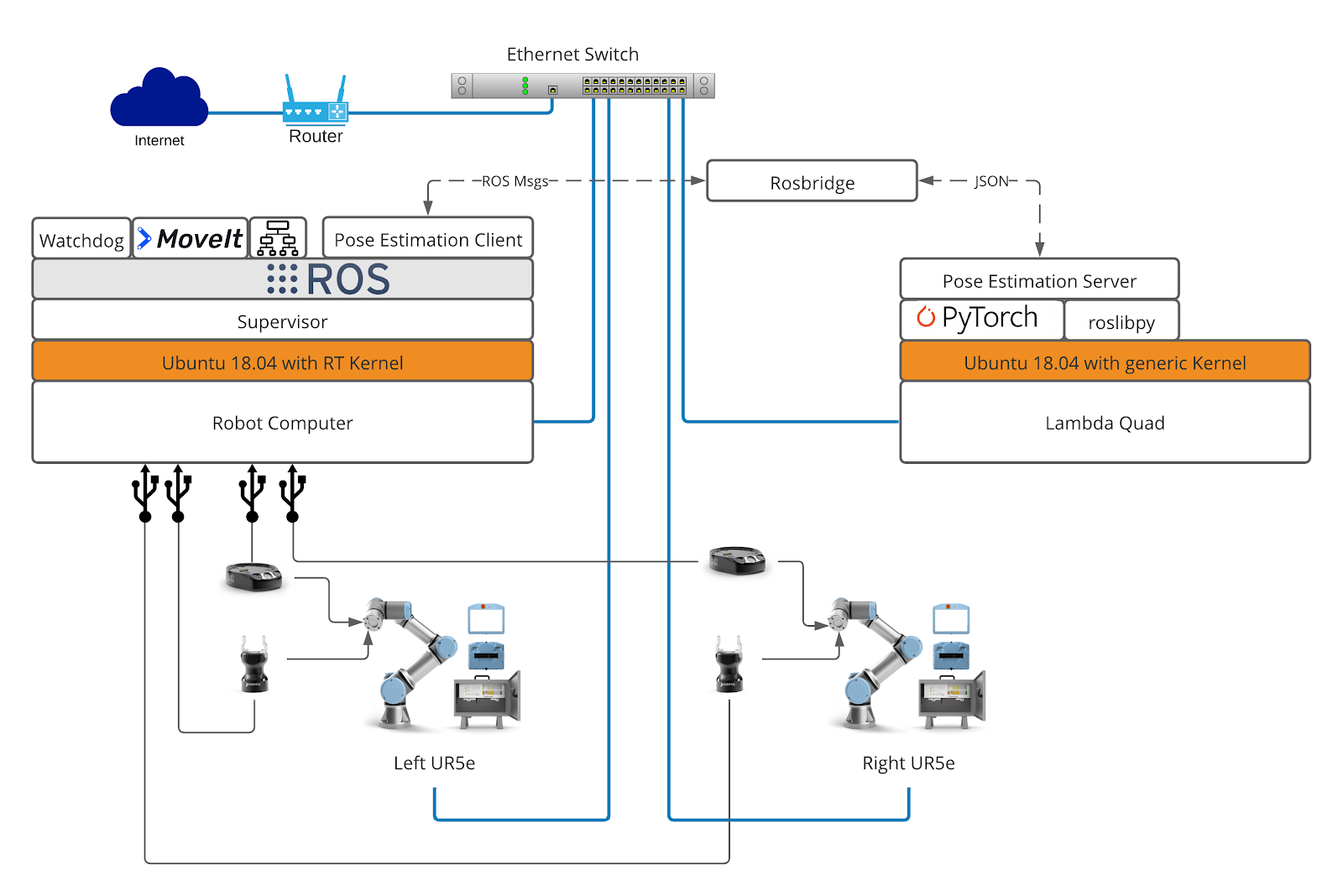}
    \caption{Our development software and hardware setup}\label{fig:hw-sw}
\end{figure}

Our system follows a heavily distributed approach at its core
(see figure \ref{fig:hw-sw}).
We have two UR5e robots, each with a control box and a Linux-based teach pendant. We use Robotiq Hand-e grippers and Robotiq wrist cameras for our end-effectors and on-hand cameras. During development, we used two computers: the ``Robot Computer'' which contains all of our system components except the deep-learning based pose estimation (vision system), and a workstation\footnote{https://lambdalabs.com/gpu-workstations/vector} for our vision system. We have this separation for two reasons. First, to enable parallel development of the vision system and the rest of the system. Second, the Linux realtime kernel on our robot computer is incompatible with NVIDIA graphics and CUDA drivers that are essential for our vision system. The robots and the computers are connected via high-frequency EthernetIP. The wrist cameras and end-effectors are connected to the Robot Computer via USB3. Connecting to the Internet was essential for remote development during COVID-related shutdowns.

We also designed and machined custom fingers for the Hand-e gripper (see figure \ref{fig:finger}).  Among other considerations, these fingers have depressions to better hold screws and stabilize the shaft, and have a beveled design to make sure that, once grasped, the elastic band does not snap out. This design came out of extensive testing, iterating on several finger designs.

\subsection{Selection of Camera Setup}

Since the physical setup of WRS competition was provided as a general guideline without fixed environmental landmarks, we chose wrist cameras over fixed cameras. The secondary reason was to account for occlusions which may occur due to the arms themselves.
We reasoned that with cameras that are eye in hand we would be better prepared to adapt while avoiding such occlusions for a wider range of environmental configuration at WRS.

Although we didn't use fixed cameras for our system, we believe that a high-quality 3D scanner, as used by team O2AS in WRC 2018 \cite{von2020team}, could augment the capability of our system.



\subsection{Task Planning with Handling of Failures}

We modeled our planning system on the classic three-tiered architecture (3T) combining deliberative planning at the upper levels with specialized planning and reactive control-flows at the lower levels \cite{kortenkamp2016robotic}. The mission planner makes commitment in plan-space to find a legal partial order for placing parts (inter-part collisions being the main heuristic) \cite{melloCorrect1991}, and the task planner (based on hierarchical task network formalism \cite{nau1999shop}) relies on a knowledge-base of domain procedures and domain objects to instantiate placement of part as a totally-ordered sequence of primitive actions. The output of these two levels is a sequence of actions and related preconditions for actuating the robot based on the taxonomy of primitive assembly skills proposed in \cite{huckaby2014knowledge}. We used behavior trees to instantiate the final plans that the task planner came up with and inject with correct physical parameters (see Section \ref{sec:bt_execution}).

We devised a classification hierarchy for planning failures that might occur in assembly planning and execution, and observed that while certain failures can be reactively handled at the lower-level, others would need to be pushed up for deliberative planning. While explaining the complete classification hierarchy is out of the scope of current paper, the most relevant classes of failures to the WRS assembly system were task planning failures relating to object grounding and non-fatal execution failures relating to probabilistic completeness. Grounding failures occur when known objects are not situated in the expected or optimal states for assembly attachments, for example the housing from Fig. \ref{fig:taskboard} being placed with the attachment end up thus a simple pick-and-place will fail, this required a revision of plan through the task-planner. An example of non-fatal execution failure is when \textit{align} action fails to find the alignment between housing and hole-feature using force-based spiral search and requires a probabilistic recovery of researching with a bigger span over task-space. We will cover our task planning in more detail in a future publication.

\subsection{Flexible Framework for Execution}
\label{sec:bt_execution}

We use a Behavior Tree-based framework for execution of tasks. We made this choice because behavior trees support high levels of reactivity and modularity, and can ultimately support flexible assembly (see section \ref{sec:framework-lesson}). For actual implementation, we develop on top of the open-source BehaviorTree.CPP\footnote{https://github.com/BehaviorTree/BehaviorTree.CPP} C++ library. It supports asynchronous and type-safe actions, and allows for creation of trees at run-time using an XML representation. It also allows for logging, replaying and live-monitoring of flow of execution through a tree. We made some changes to the library source code to better suit it for our needs. We chose this library because it is well-maintained, well-documented, and has an active circle of users. Notably, the Navigation2 project is also built upon this library \cite{macenski2020marathon2}. See section \ref{sec:imp-btf} for implementation details.

\subsection{Watchdog Subsystem}

It remains a fact that for any complex system the components of are not 100\% reliable, including ours. Even before running experiments to determine reliability quantitatively, during development and software testing, we observed occasional failures (see section \ref{sec:obstacle}). Some components failed more often than others, and some failures were more critical than others. While many of these failures were fixed by patching software bugs, some were more tricky and non-deterministic (see section \ref{sec:obstacle} for examples). It is, in theory, possible to fix these issues by isolating their root cause and repairing them. In practice, however, it is often impractical because we might not have access to the buggy code, or fixing such bugs would be too costly in terms of developer time. It is easier to diagnose system-level failures, and then have a watchdog system that can simply ``restart'' failing system components as they fail\cite{wang2019robotic}. Thus, our watchdog subsystem did three simple things:

\begin{enumerate}
    \item Subscribe to periodic ``heartbeats'' from all component nodes/processes (robots, grippers, cameras etc.).
    \item Monitor log file for communication failures.
    \item Restart faulty nodes.
\end{enumerate}

\subsection{Vision and Learning to Handle Specular Objects}

Objects in WRC 2020 differ from those used in recent work on 6D pose estimation \cite{li2018deepim} \cite{kehl2017ssd} \cite{park2019pix2pose}. In particular, most of the WRC 2020 objects lack texture, have varied symmetry, high specular reflection, radial anisotropy, and have limited features. We approached the problem of vision in a strong departure from merely task-specific vision \cite{eppner2016lessons}, focusing on solving the more general problem of estimating the 6D pose of textureless, symmetric objects in a given workspace using RGB vision. See our vision paper \cite{hu2020pose} for more details.

\begin{figure}
    \centering
    \includegraphics[width=0.3\textwidth]{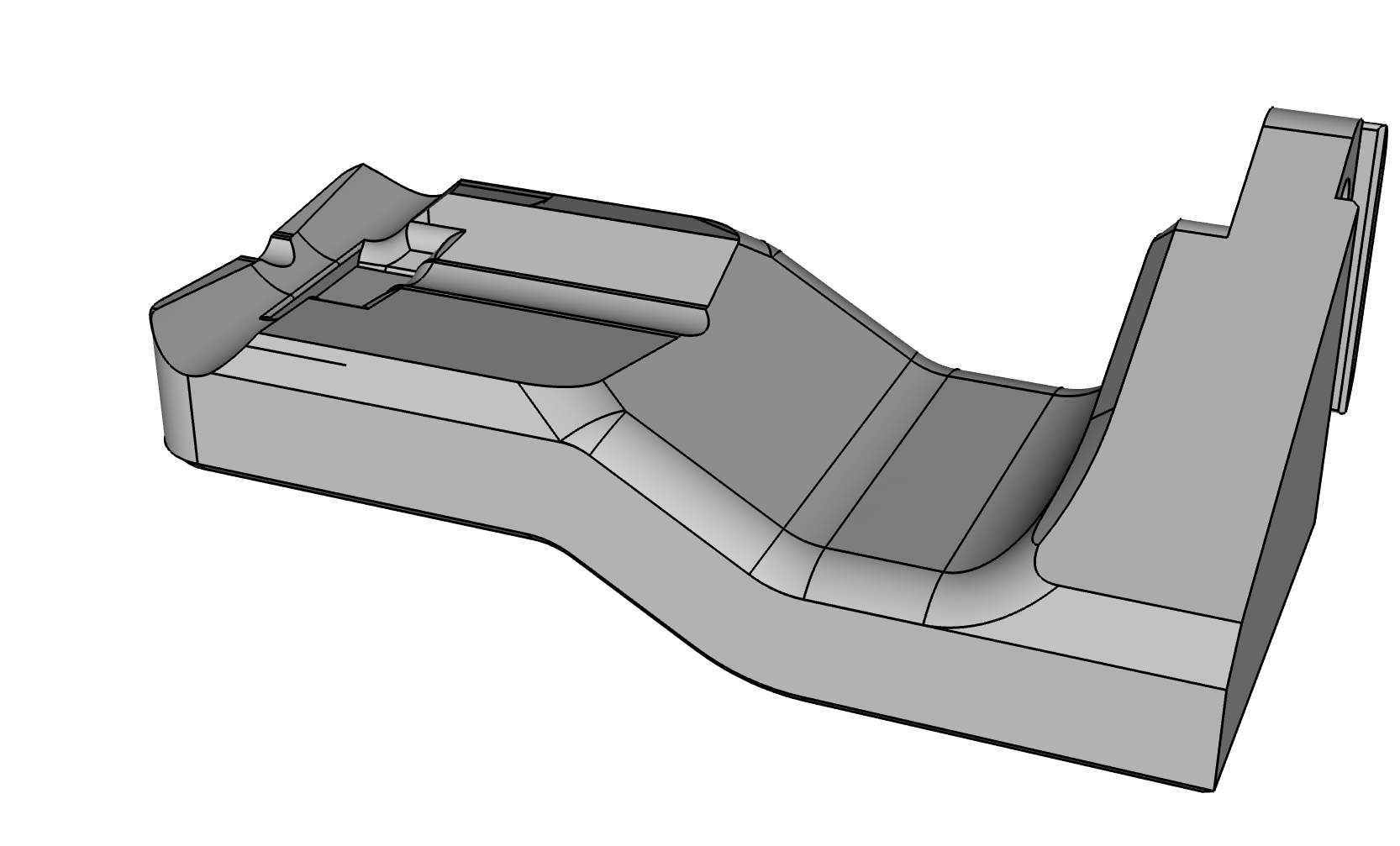}
    \caption{Our finger design.}
    \label{fig:finger}
\end{figure}

\section{Actual Implementation}\label{sec:impl}

\begin{table*}[]
\centering
\begin{tabular}{@{}p{0.08\linewidth}p{0.05\linewidth}p{0.35\linewidth}p{0.13\linewidth}p{0.13\linewidth}p{0.13\linewidth}@{}}

\toprule
\textbf{Name}    & \textbf{Type} & \textbf{Function}                                                                                                                                                                                                                                                                                                                                                                                                                          & \textbf{RUNNING}                                     & \textbf{SUCCESS}                            & \textbf{FAILURE}                               \\ \midrule
\textit{Sequence}     & Control Flow       & Ticks its children from left to right.                                                                                                                                                                                                                                                                                                                                                                                                     & While a child node is RUNNING                                & When all its children return SUCCESS                & As soon as a child returns FAILURE                     \\ \midrule
\textit{Fallback}     & Control Flow       & Ticks its children from left to right.                                                                                                                                                                                                                                                                                                                                                                                                     & While a child node is RUNNING                                & As soon as a child returns SUCCESS                  & When all its children return FAILURE                   \\ \midrule
\textit{Parallel}     & Control Flow       & Ticks all its children concurrently.                                                                                                                                                                                                                                                                                                                                                                                                       & While a child node is RUNNING                                & When $k$ (parameter) children return SUCCESS        & When more than $k$ (parameter) children return FAILURE \\ \midrule
\textit{MoveJoint}    & Action             & Accepts either a precomputed keyframe name or numeric joint values as a target (joint goal). If possible, moves robot arm to the joint goal                                                                                                                                                                                                                                                                                                & While planning trajectory or moving to the goal              & When joint goal is reached                          & Trajectory cannot be planned or executed               \\ \midrule
\textit{MoveEE}       & Action             & Stands for ``move end-effector''. Accepts a 6D cartesian space goal for the end-effector. Performs inverse kinematics calculations and moves the robot arm so that the end-effector frame aligns with the goal frame.                                                                                                                                                                                                                      & While planning trajectory or moving to the goal              & When goal is reached                                & Trajectory cannot be planned or executed               \\ \midrule
\textit{Grasp}        & Action             & Opens/Closes the gripper fingers. Optionally also accepts a real value between 0 and 1 for partial closure.                                                                                                                                                                                                                                                                                                                                      & While opening/closing gripper                                    & Desired gripper goal is reached                     & Gripper cannot be closed/opened                               \\ \midrule
\textit{MoveUntilFF}  & Action             & Stands for ``move until force feedback''. Takes in a wrench (3D force and torque) threshold as argument. Moves the robot along the current axis through the robot fingers until the UR5e's force sensor detects a force greater than the specified threshold.                                                                                                                                                                              & While the robot is moving                                    & Detected force greater than threshold               & Never. This action is used with a Timeout decorator    \\ \midrule
\textit{SearchAlign}  & Action             & Performs a spiral search on the current plane (perpendicular to the gripper fingers) for a hole or a cavity                                                                                                                                                                                                                                                                                                                                & While the robot is still searching                           & When hole/cavity is detected                        & Never. This action is used with a Timeout decorator    \\ \midrule
\textit{NJInsert}     & Action             & Stands for ``non-jamming insert''. After finding a hole (using SearchAlign), if we simply push the object in the robot's hand into the hole, it is very likely that it will get jammed (because of the strict allowances). Thus, we also push the object at a slight angle in the opposite direction from where the sensor feels the greatest force. The direction is changed in real time in response to the current force sensor values. & While the insertion is in progress                           & Detected force greater than threshold               & Never. This action is used with a Timeout decorator    \\ \midrule
\textit{EstimatePose} & Action             & Given RGB image and name of object, estimates the 6D pose of object(s) in the camera frame.                                                                                                                                                                                                                                                                                                                                  & Until the request is processed by the pose estimation server & If at least one object with the given name is found & No objects with the given name are found               \\ \midrule
\textit{ComputeGrasp} & Action             & Given name of object and its 6D pose (in any valid coordinate frame, usually the camera frame), computes the pose of the end-effector to grasp the object in a stable way. This grasp pose is precomputed for all objects using GraspIt \cite{miller2004graspit} and stored in a database.                                                                                                                                                 & Until request is processed                                   & When grasp pose is computed                         & If object name is not found in database                \\ \midrule
Misc. Conditions      & Condition          & Checks the Blackboard for conditions. It is the responsibility of various action nodes to update the Blackboard appropriately.                                                                                                                                                                                                                                                                                                             & Never                                                        & If condition is true                                & If condition is false                                  \\ \bottomrule
\end{tabular}
\caption{Nodes in our BT Framework.}
\label{tab:bt-fw}
\end{table*}
\subsection{Behavior Tree-based Framework}\label{sec:imp-btf}

A behavior tree has two kinds of nodes: control flow nodes, which are internal nodes of the tree, and execution nodes, which are leaf nodes of the tree. An execution signal called a ``tick'' flows from from the root of the tree top-to-bottom, left-to-right at a certain frequency ($\approx1000$Hz in our case). There are four main kinds of control flow nodes: Sequence (denoted by $\rightarrow$), Fallback (denoted by $?$), Parallel (denoted by $\rightrightarrows$), and Decorator (diamond-shaped node). There are two kinds of execution nodes: Action and Condition.

For any behavior tree based framework, the execution nodes are domain-specific ``behaviors'' (Action nodes) or assertions about the environment state (Condition nodes). Control flow nodes are usually common across domains.

For our initial set of behaviors (execution nodes), we relied on a taxonomy of primitive actions for industrial assembly \cite{huckaby2014knowledge}. We added a few more behaviors to our execution framework while development and testing.

All nodes can return one of three states--RUNNING, SUCCESS, and FAILURE--when they are ``ticked'' (ticks flow into them). We have described the behavior tree nodes used in our execution framework in table \ref{tab:bt-fw}. We also use some more advanced control nodes in our execution (like ReactiveSequence), but they are not relevant for this work. We encourage the reader to look the additional nodes in Behavior Trees in Robotics and AI \cite{colledanchise2018behavior}.

We refer to the program that loads Behavior Trees at runtime, and executes them, as the BT executor. A crucial point to note here is that the BT executor runs in a different Supervisor process group than the robot hardware. Thus, when we restart a hardware node or even the entire robot system, the behavior tree keeps running, and can resume execution--this greatly improves our fault-tolerance. 

With every execution instance, we have a key-value store, called the \emph{Blackboard}, that can be used to maintain state and for communication across nodes.

\begin{figure*}
    \centering
     \begin{subfigure}[b]{0.32\textwidth}
         \centering
         \includegraphics[width=\textwidth]{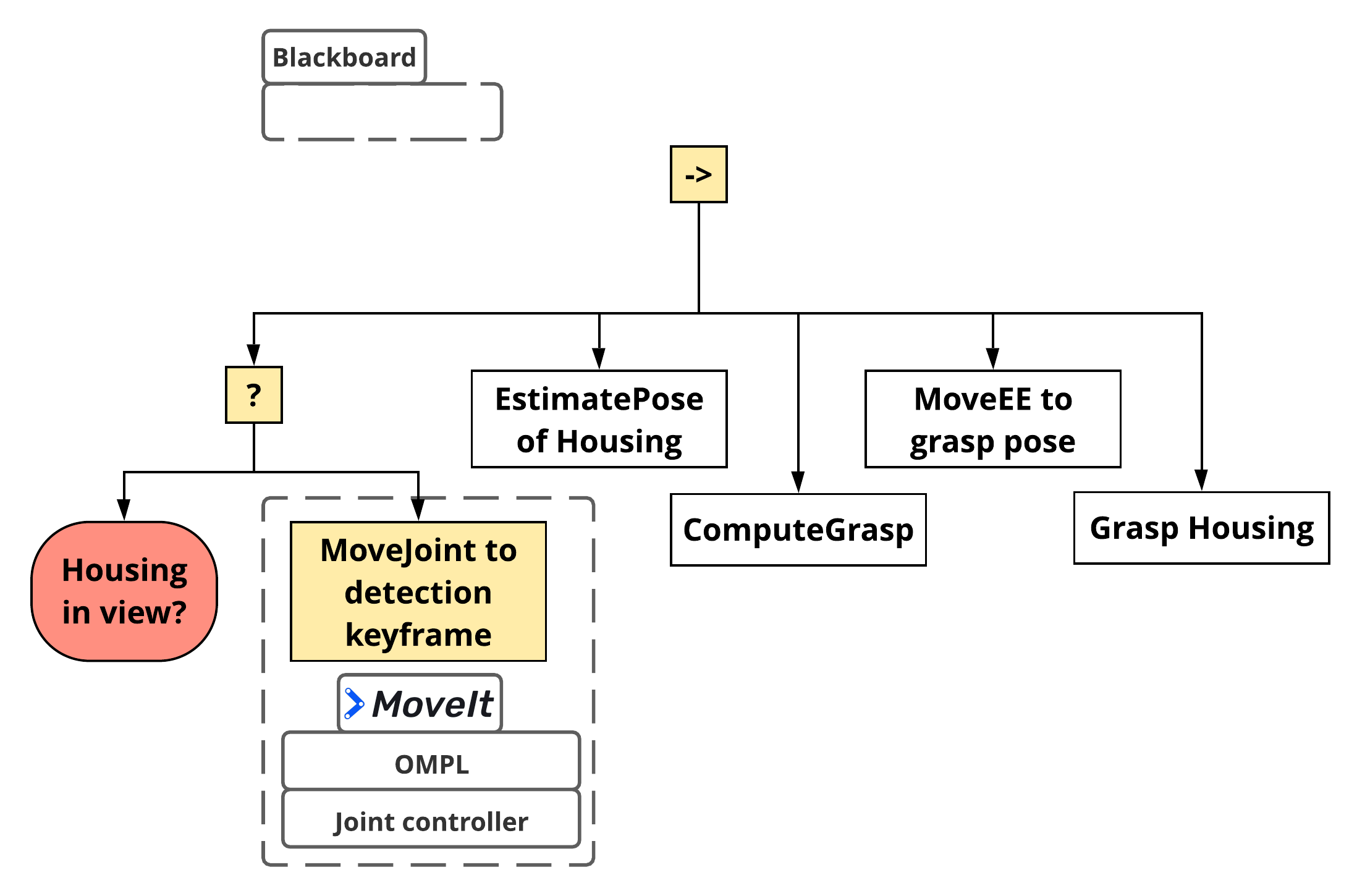}
         \caption{The executor finds that the housing is not in camera view. Moves the robot to a keyframe from where it can look for housing.}
         \label{fig:exec1}
     \end{subfigure}
     \hfill
     \begin{subfigure}[b]{0.32\textwidth}
         \centering
         \includegraphics[width=\textwidth]{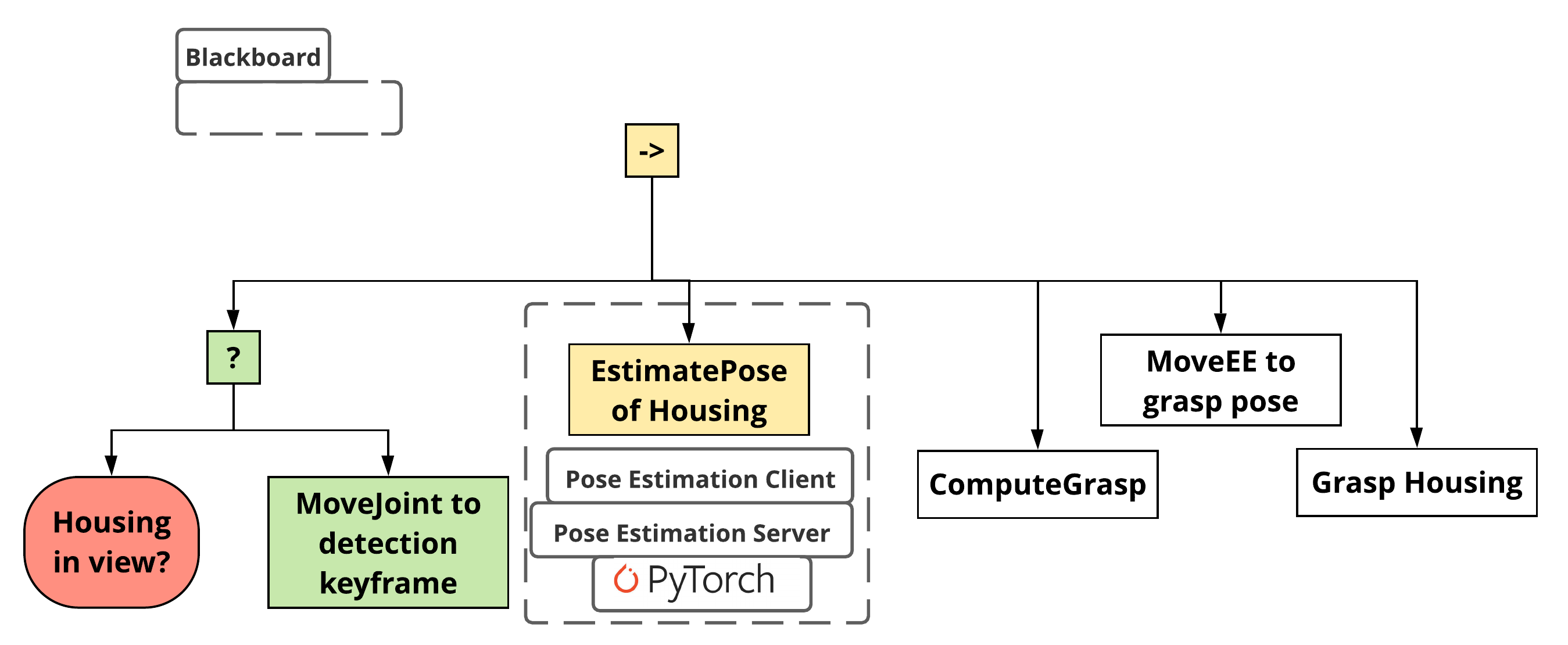}
         \caption{Move to keyframe is complete. Executor calls pose estimation stack.}
         \label{fig:exec2}
     \end{subfigure}
     \hfill
     \begin{subfigure}[b]{0.32\textwidth}
         \centering
         \includegraphics[width=\textwidth]{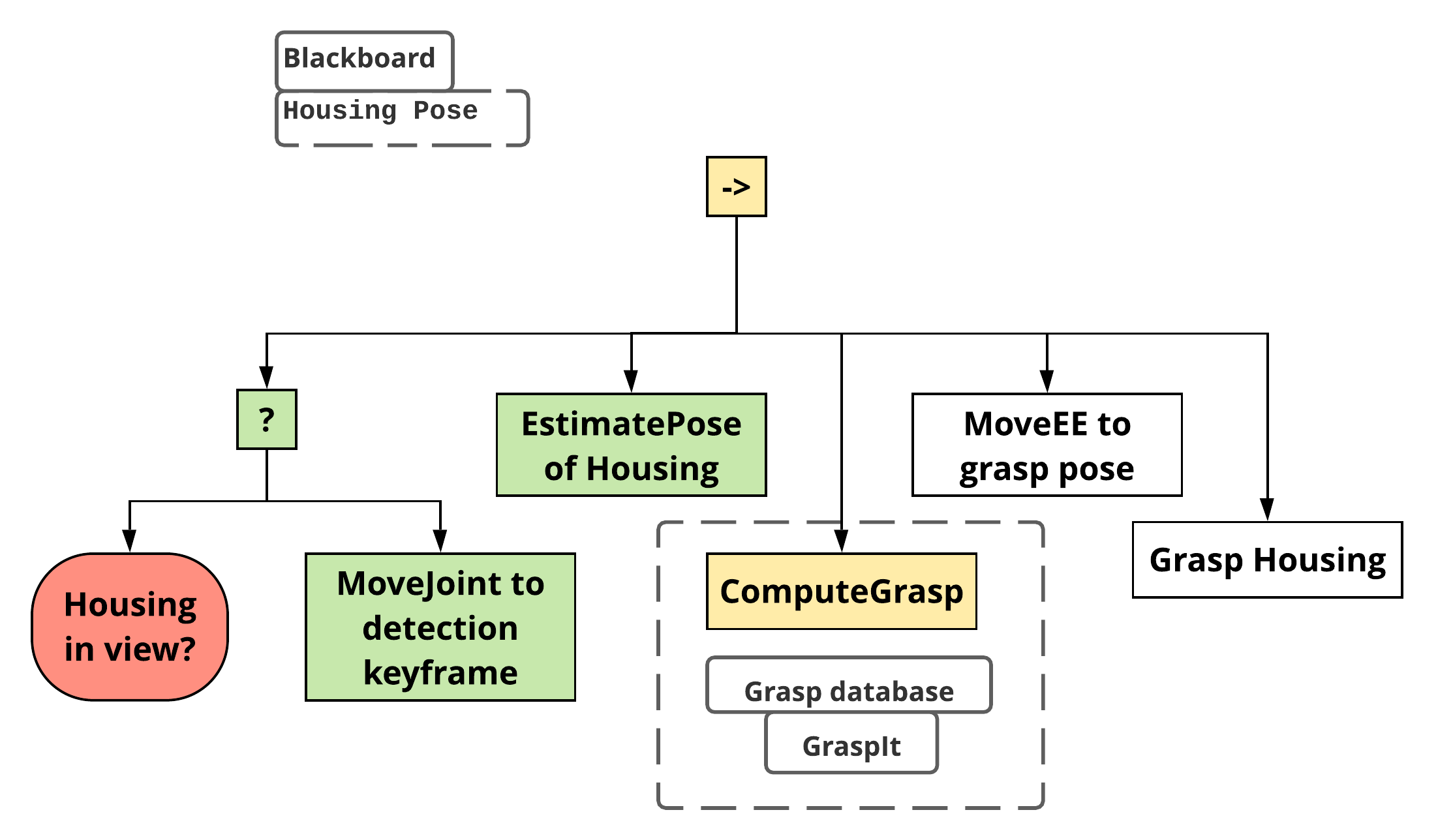}
         \caption{Pose is detected and written to Blackboard. Executor calls for computing grasp.}
         \label{fig:exec3}
     \end{subfigure}
     \hfill
     \begin{subfigure}[b]{0.32\textwidth}
         \centering
         \includegraphics[width=\textwidth]{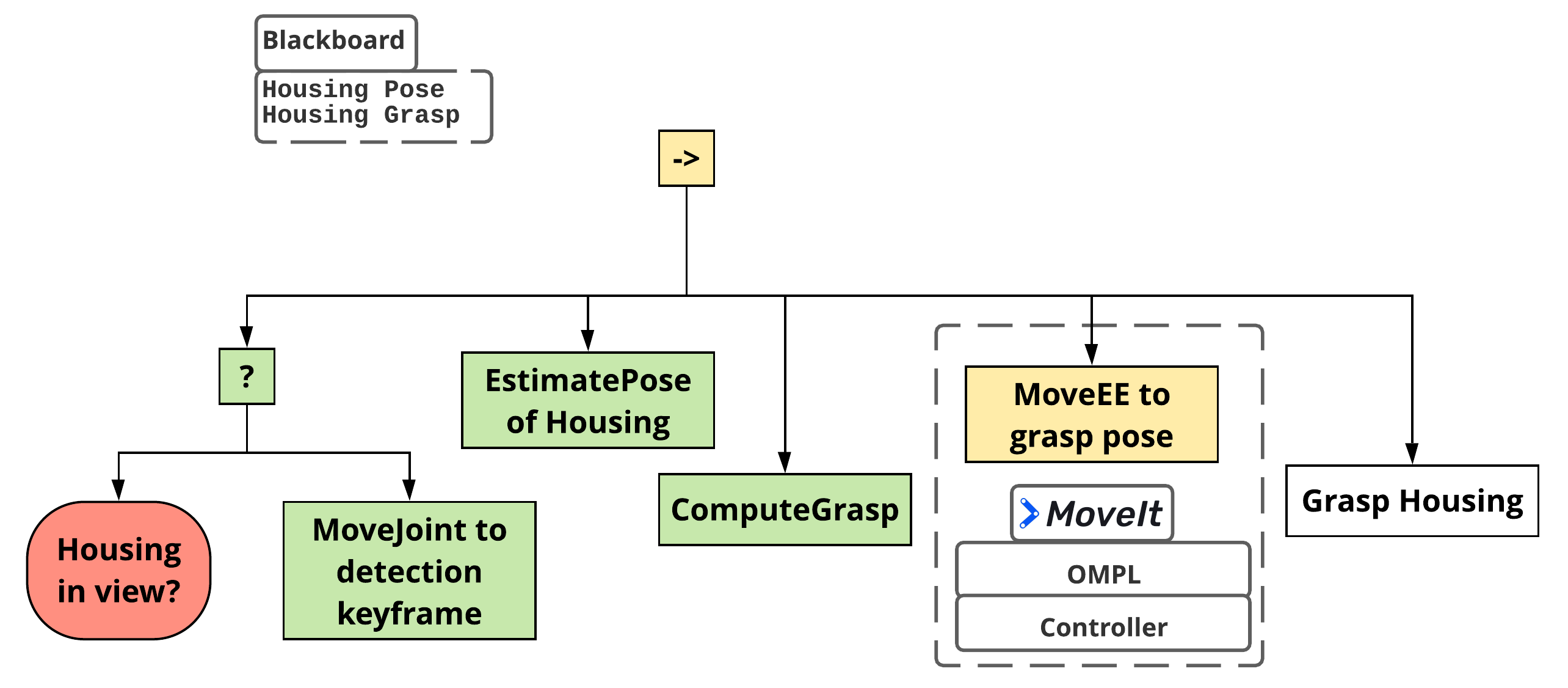}
         \caption{Grasp is computed and written to Blackboard. Executor runs \textit{MoveEE} for moving to grasp pose.}
         \label{fig:exec4}
     \end{subfigure}
     \hfill
     \begin{subfigure}[b]{0.32\textwidth}
         \centering
         \includegraphics[width=\textwidth]{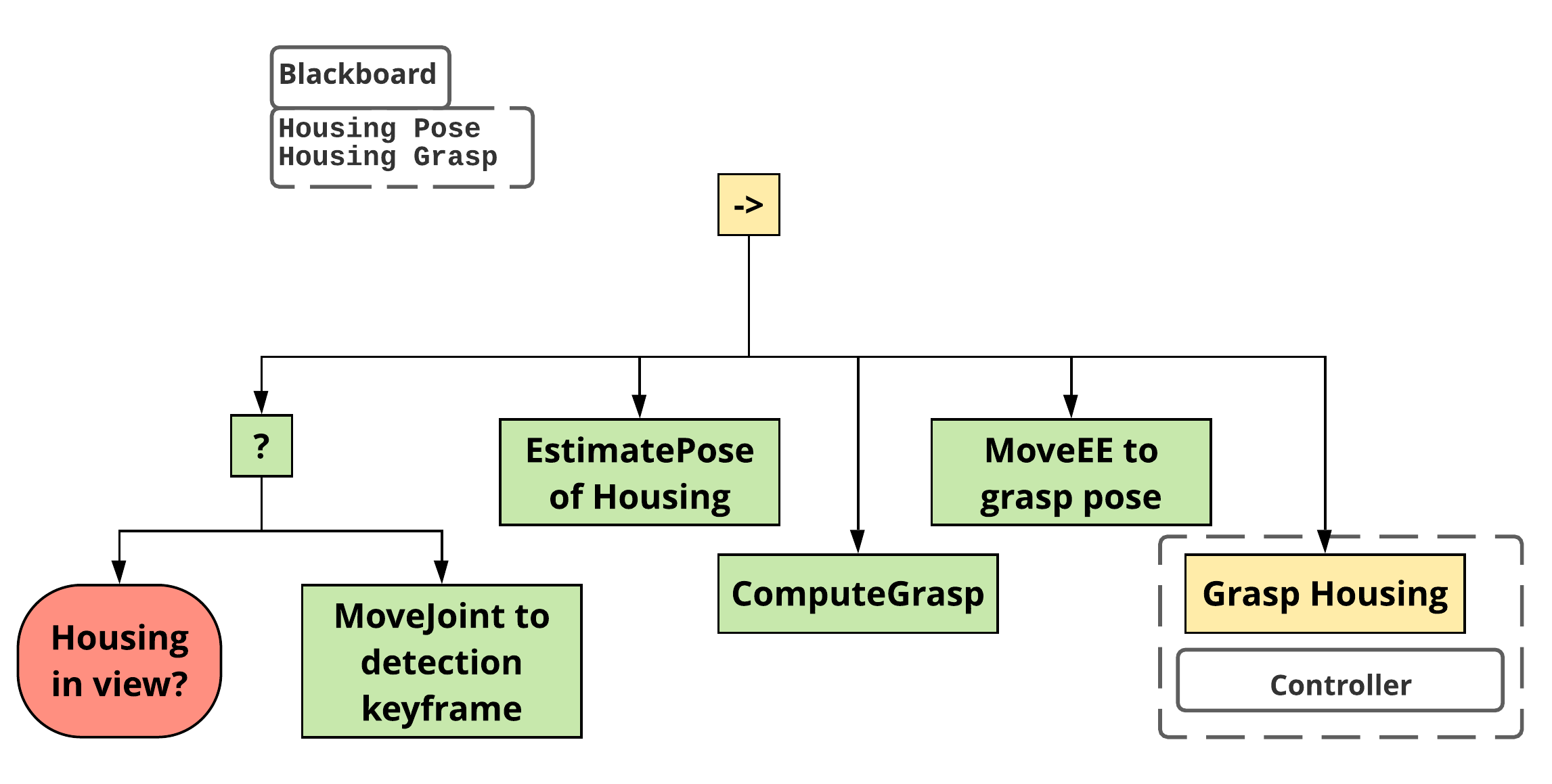}
         
         \caption{Robot has moved to the grasp pose. Executor calls to close the gripper.}
         \label{fig:exec5}
     \end{subfigure}
     \hfill
     \begin{subfigure}[b]{0.32\textwidth}
         \centering
         \includegraphics[width=\textwidth]{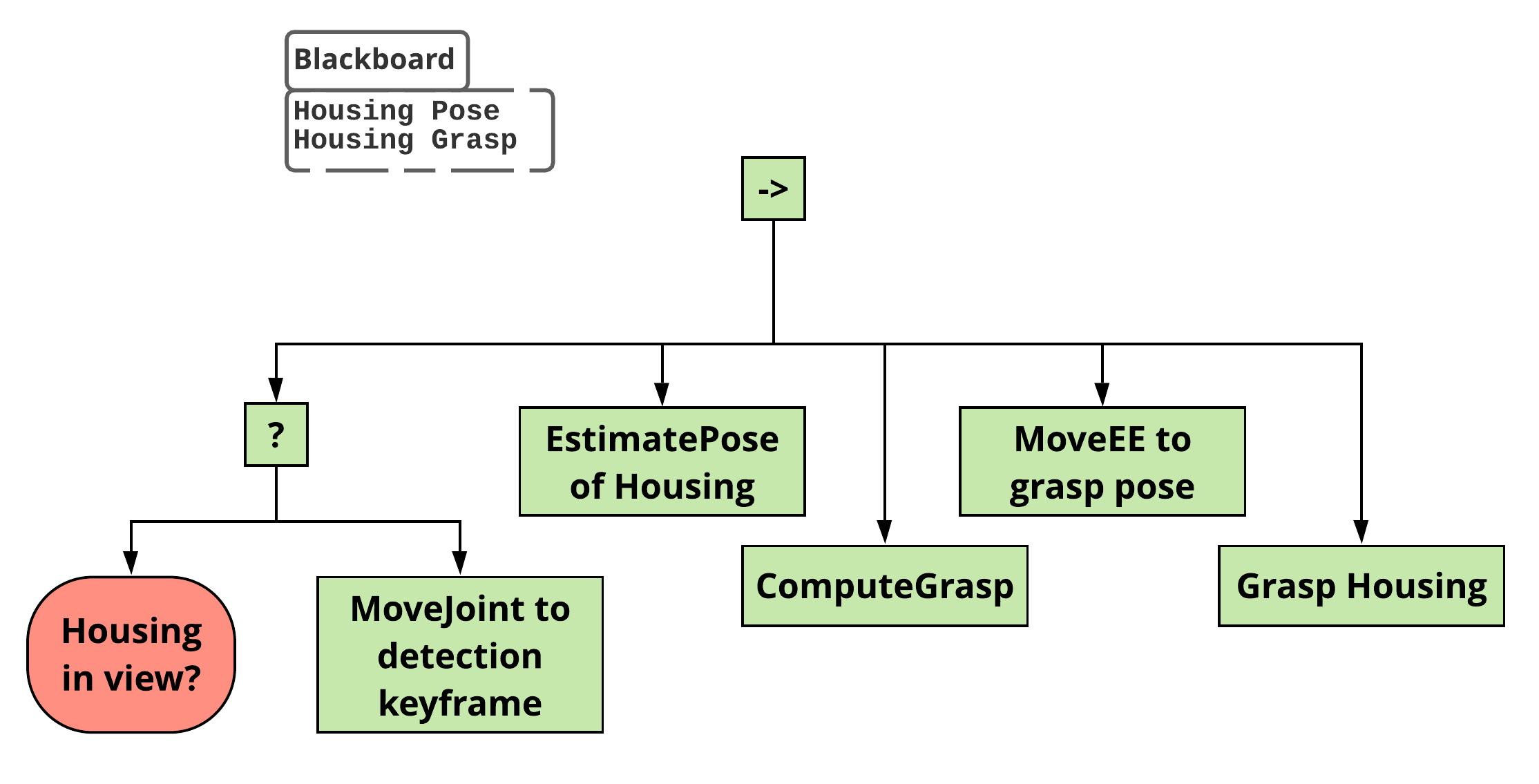}
         \caption{Gripper is closed, housing is grasped. Execution complete.}
         \label{fig:exec6}
     \end{subfigure}
        \caption{BT execution for finding and picking up housing. Color code: yellow (RUNNING), red (FAILURE), green (SUCCESS).}
    \label{fig:execution}
\end{figure*}

\begin{figure}
    \centering
    \includegraphics[width=0.44\textwidth]{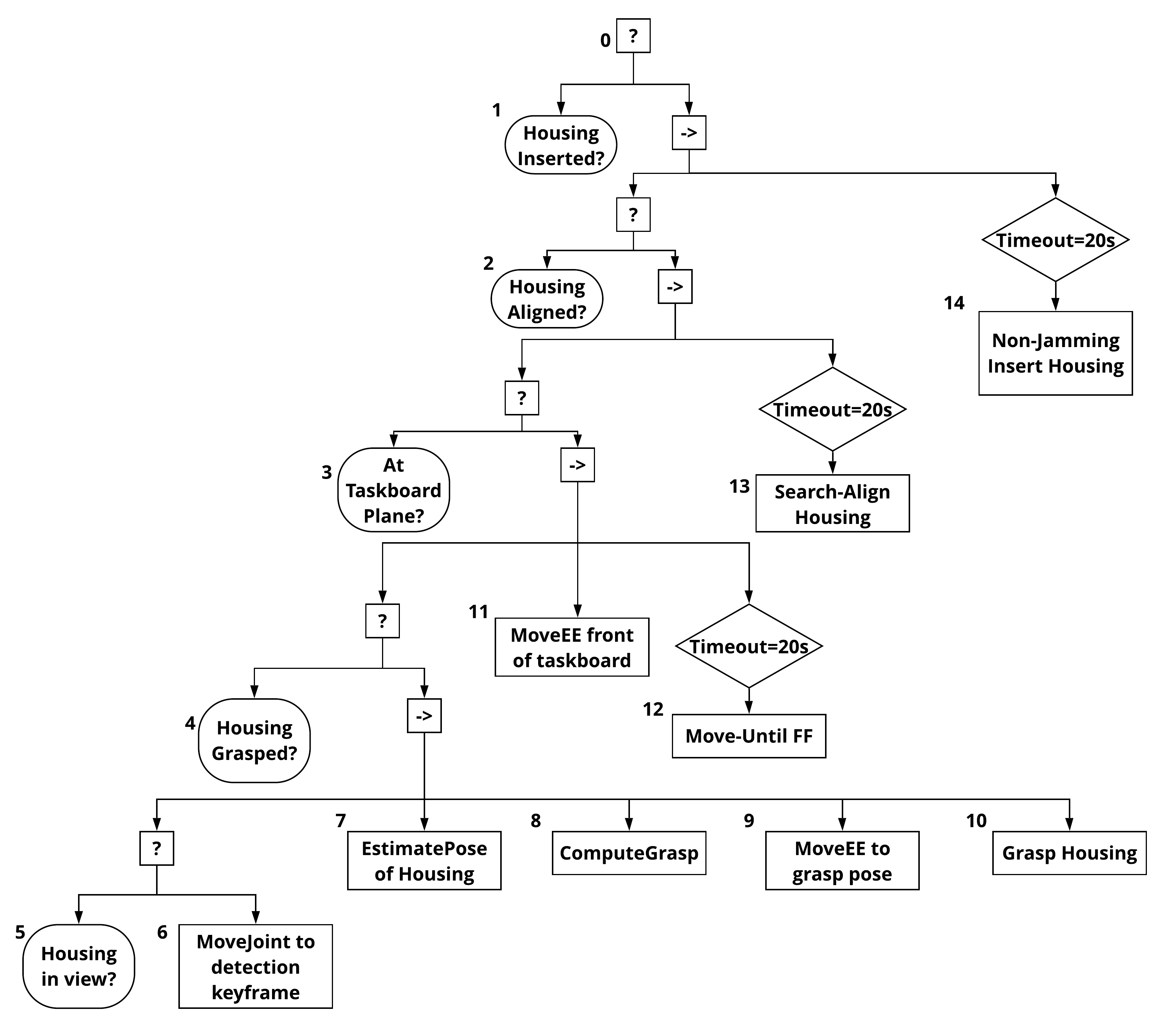}
    \caption{Behavior tree to insert housing. Some nodes are abstractions to enable easier understanding. Nodes are numbered in the order they are ticked. Note that the subtree with nodes numbered 5 through 10 can be used to find and pick up the housing. This demonstrates modularity and reusability in behavior trees.}
    \label{fig:insert-housing}
\end{figure}

\subsection{Execution Walkthrough}

We show an example behavior tree that finds, picks up, and inserts an housing into the taskboard in figure \ref{fig:insert-housing}. This behavior tree can be written and executed on our framework. To demonstrate how exactly the execution proceeds and to show the processes behind the scenes, we walk through the subtree for finding and picking up the housing (nodes 5 through 10) in figure \ref{fig:execution}.

\subsection{Use of Standard Software Packages}\label{sec:std-pkgs}

Obviously, any complex software system would be prohibitively expensive if built entirely from scratch. In this section we talk the main standard packages that we used.

\subsubsection{ROS} Robot Operating System (ROS) is a open-source middleware that allows robotics projects to build on top of a common infrastructure \cite{roscore} providing
access to more than 3000 user-contributed packages. We used ROS Melodic installed natively on Ubuntu 18.04 machines.

\subsubsection{BehaviorTree.CPP and Groot} BehaviorTree.CPP\footnote{https://github.com/BehaviorTree/BehaviorTree.CPP} is an open-source C++ library. It supports asynchronous and type-safe actions, and allows for creation of trees at run-time using an XML representation. Groot is a GUI tool that allows for creating and monitoring behavior trees using a graphical interface. We used version 3.5.3 of BehaviorTree.CPP and 1.0.0 of Groot.

\subsubsection{MoveIt} MoveIt is a robot motion planning framework that provides a unified interface to trajectory execution, forward and inverse kinematics, collision detection and planning algorithms for a large vareity of robots \cite{coleman2014reducing}. We used MoveIt's Melodic version 1.0.7.

\subsubsection{OMPL and LazyPRM} We use sampling-based motion planning algorithms for our robot arms, implemented in Open Motion Planning Library (OMPL) \cite{sucan2012open}. Through experimentation, we found that the multi-query LazyPRM motion planner \cite{bohlin2000path} worked best for most cases. This is likely because our robot arms operated in the same physical space during the execution. We used OMPL version 1.5.1.

\subsubsection{PyTorch} PyTorch is an optimized tensor library for deep learning \cite{paszke2019pytorch}. We used PyTorch write and train our pose estimation models. We used PyTorch version 1.5.1.

\subsection{Codebase Details}

Excluding the vision system, most of our codebase was written in C++. We used the GCC C++ 2014 standard compliant compiler. We make heavy use of RAII-style \cite{stroustrup1994design} ``modern C++'' primitives like smart pointers and locking wrappers to avoid memory leaks and deadlocks. For our build system, we use Catkin\footnote{https://github.com/ros/catkin}, which is a CMake-based build system. It is also the official build system for ROS-1. Our vision system used PyTorch and was almost entirely written in Python (3.6). Our behavior trees and some ROS config files are written in XML. We also use YAML for config files. See our codebase statistics in table \ref{tab:codebase-stats}. We report statistics for our entire codebase (all), which includes the source code of packages we built from source, and for the code written entirely by us (ours). 
We used open-source software scc\footnote{https://github.com/boyter/scc} to compute our codebase statistics. Code complexity is cyclomatic complexity \cite{watson1996structured}.

\begin{table*}[]
\centering
\begin{tabular}{@{}lllllll@{}}
\toprule
Language     & Files (all) & Lines (all) & Code complexity (all) & Files (ours) & Lines (ours) & Code complexity (ours) \\ \midrule
C++          & 680         & 182181    & 17684                 & 81           & 13743      & 1516                   \\
C/C++ header & 635         & 108608    & 2805                  & 66           & 4625       & 60                     \\
Python       & 112         & 18132     & 1011                  & 61           & 11855      & 681                    \\
XML          & 287         & 7953      & 0                     & 65           & 4680       & 0                      \\
CMake        & 168         & 10975     & 190                   & 21           & 2366       & 18                     \\
YAML         & 78          & 3362      & 0                     & 20           & 674        & 0                      \\
\midrule
Total        & 1960        & 331211    & 21690                 & 314          & 37943      & 2275                   \\
\bottomrule
\end{tabular}
\caption{Codebase language statistics}\label{tab:codebase-stats}
\end{table*}


\section{Performance}\label{sec:perf}
\subsection{Reliable Execution}

In this work, we focus on overall system reliability. For good end-to-end execution during the competition, it would be important that everything else run reliably and correctly for an extended period of time (at least an hour) without any intervention. Thus, we designed a task that used all our primitive skills (except those relating to insertion, namely MoveUntilFF, SearchAlign and NonJammingInsert) and which could run in a loop. We measured the mean time until intervention (MTUI) on this task. A video of this task and automatic recovery of failures can be found by following the provided link\footnote{https://bit.ly/35w6Ndw}.


We found that without any special reliability measures our system's MTUI was very low. This was primarily because the sampling-based motion planners didn't always find a valid solution (within the time limit), and many of the open-source hardware drivers provided unreliable interfaces. See section \ref{sec:obstacle} for more details. Enabling re-execution and recovery using behavior tree reactivity improved the MTUI a little. Finally, adding in the watchdog subsystem that monitored all hardware components and could even restart the entire system in the worst kinds of failure improved the MTUI immensely. That is, using the watchdog subsystem to restart failed components and using behavior trees to resume execution and recover in the face of failure pushed our system's MTUI to more than several hours.

With only reactive recovery using Behavior Trees (high-level), we ran the task 20 times. We observed that the MTUI was 7m 12s with standard deviation 6m 35s. With our watchdog subsystem enabled (and BT reactivity), although we ran the looped task for close to three hours on multiple occasions, the system always managed to recover on its own (without intervention). Thus we could not calculate MTUI. But we can be fairly confident that the MTUI is greater than three hours. This is a massive boost.

At the time of writing this paper, we could not get our insertion actions to run reliably because of a hidden source of noise that we couldn't identify despite great efforts. Only a few days before the robots were due for return to our sponsors did we discover that a nearby WiFi Access Point was interfering with the force-torque sensors of the UR5e robots. We leave performance evaluation of our end-to-end execution for the final competition or perhaps, in another work.

\subsection{Obstacles to Performance} \label{sec:obstacle}

Here we describe the various hardware and software issues that prevented the system from performing robustly, and what measures we took to mitigate them. Our main objective here is to provide a guide for future robot system designers to quickly quell common issues.

\subsection{Hardware Issues}
\label{sec:hardware_issues}

\begin{itemize}
    \item In the hardware design is it essential to consider the distribution of load across controllers. Due to hardware overload (too many cameras on one controller) we saw frequent driver failures for the cameras. This issue also overflowed to grippers which were on the same controller as the camera. Fixed by connecting the cameras to different USB controllers and grippers to a third bus in the computer.
    \item Interference between USB-2 and USB-3 protocols/cables caused frequent driver failures for, both, the cameras and the grippers. Fixed by using USB 3.0 extenders.
    \item Gripper disconnection and device aliasing issues. Despite grippers being on different USB controller we still had to restart grippers occasionally using the watchdog.
    \item Robot communication failure. The protocol for ROS/UR-5e communication has occasional failures, which requires a restart. Detected from the log file. 
    \item Robot oscillatory behavior when compliance controller was turned on. It took months before we identified the root cause: a WiFi access point nearby caused the robot's force-torque sensor to be noisy.
\end{itemize}

\subsection{Software Issues}

\begin{itemize}
    \item ROS controllers failing to start. Fixed by monitoring a list of active controllers and restarting failed controllers.
    \item Stochastic motion planners will sometimes fail even when a feasible plan exists. First-order recovery is based on trying a simple replan.
    \item Race condition on robot description (robot description not loaded to ROS parameter server before needed by Compliance controller). Fixed by adding a 5 second wait before starting the compliance controller.
\end{itemize}




\section{Lessons Learned}\label{sec:lesson}


\subsection{A Framework Focused on Fault-recovery, Modularity, Reactivity Essential for Flexible Assembly}\label{sec:framework-lesson}

We would like to highlight that our framework encodes all the principles necessary for a ``level 5'' production system \cite{wrs2020rules}. With a repertoire of well-tested behaviors, and an automatic way to compose them into tasks to accomplish a mission, a zero-day changeover is much more realistic. With transparent system-level recovery, and well-defined and reactive mission-level and task-level recovery, the operation rate of a production system can be greatly enhanced. Further enhancement could be made in this aspect by failure prediction using continual learning methods. Finally, our finger design and vision system provide an early template for leanness and reusability by facilitating jigless, flexible assembly. 


\subsection{Good Embodiment can Reduce Complexity}

We note that making use of intelligent hardware design (embodiment) can significantly reduce the complexity of the overall system. This is clearly illustrated with our finger design (figure \ref{fig:finger}). Such a finger design made grasping simple. For example, even if the gripper started grasping the pulley at a slight offset from the optimal grasp, the pulley would just ``fall into place'' because of the depressions.
We would also like emphasize the inverse, that is, suboptimal hardware design decisions can blow up the complexity of the system. For example, Robotiq wrist cameras were off-center from the gripper axis (looked down at an angle from the wrist), which introduced several challenges for the vision system.

\subsection{ROS 1 Falls Short for Industrial Scale}

While ROS 1 helped us immensely as a basis for our system, we believe that it falls short on many fronts for developing an assembly system at industrial scale. 

\begin{enumerate}
    \item Integrating the two robot arms and grippers into an integrated system was challenging and required intricate namespace management. ROS 1 is not well-suited for multi-robot settings.
    \item ROS 1 does not natively support real-time control. Without true real-time control, it is not possible to make formal guarantees.
    \item Limited intermediate and advanced tutorials/guides. No standard guidelines on how to structure a complex robot system.
\end{enumerate}

We realize that the developers of ROS (and ROS 2) and ROS Industrial\footnote{https://rosindustrial.org/} recognize these issues and are trying to address them in ROS 2. We are hopeful about future robot system support at industrial scale.

\section*{Acknowledgment}

The authors would like to thank Nahid Sidki, Omar Aldughayem, and Hussam Alzahrani from RPD Innovations. We would also like to thank Jack Griffin for his help with the finger designs.



%




\bibliographystyle{IEEEtran}
\bibliography{main}

\end{document}